\documentclass{article}

\PassOptionsToPackage{numbers,sort&compress}{natbib}
\usepackage[main, final]{neurips_2026}

\usepackage[utf8]{inputenc}
\usepackage[T1]{fontenc}
\usepackage{hyperref}
\usepackage{url}
\usepackage{booktabs}
\usepackage{colortbl}
\usepackage{multirow}
\usepackage{makecell}
\usepackage{amsfonts}
\usepackage{amsmath}
\usepackage{amssymb}
\usepackage{nicefrac}
\usepackage{microtype}
\usepackage{xcolor}
\usepackage{graphicx}
\usepackage{wrapfig}
\usepackage{enumitem}
\usepackage{caption}
\usepackage{float}
\usepackage{tabularx}
\usepackage{siunitx}
\usepackage{eccvabbrv}   

\definecolor{linkcolor}{RGB}{80,135,242}
\hypersetup{
  colorlinks=true,
  linkcolor=linkcolor,
  citecolor=linkcolor,
  urlcolor=linkcolor,
}

\usepackage[capitalize]{cleveref}
\crefname{section}{Sec.}{Secs.}
\Crefname{section}{Section}{Sections}
\crefname{table}{Tab.}{Tabs.}
\Crefname{table}{Table}{Tables}

\definecolor{mygreen}{RGB}{0,160,10}
\providecommand{\ns}[1]{}            
\providecommand{\keywords}[1]{}      

\makeatletter
\renewcommand{\@noticestring}{}
\makeatother

\title{Semantic-Aware, Physics-Informed, Geometry-Grounded Weather Video Synthesis}

\author{%
  \textbf{Chenghao Qian}\textsuperscript{1} \quad
  \textbf{Nedko Savov}\textsuperscript{2} \quad
  \textbf{Lingdong Kong}\textsuperscript{3} \quad
  \textbf{Yeying Jin}\textsuperscript{3} \quad
  \textbf{Rui Song}\textsuperscript{4} \\[3pt]
  \textbf{Wenjing Li}\textsuperscript{1,5}\thanks{Corresponding authors.} \quad
  \textbf{Zhun Zhong}\textsuperscript{5}\footnotemark[1] \quad
  \textbf{Jiaqi Ma}\textsuperscript{4} \quad
  \textbf{Gustav Markkula}\textsuperscript{1} \quad
  \textbf{Luc Van Gool}\textsuperscript{2} \\[6pt]
  \normalfont
  \textsuperscript{1}University of Leeds, UK \quad
  \textsuperscript{2}INSAIT, Sofia University ``St.~Kliment Ohridski'' \\
  \textsuperscript{3}National University of Singapore, Singapore \quad
  \textsuperscript{4}University of California, Los Angeles, USA \\
  \textsuperscript{5}Hefei University of Technology, China \\[5pt]
  Project page: \url{https://jumponthemoon.github.io/w-crafter/}
}

\begin{document}

\maketitle

\begin{figure}[H]
    \centering
    \includegraphics[width=\linewidth]{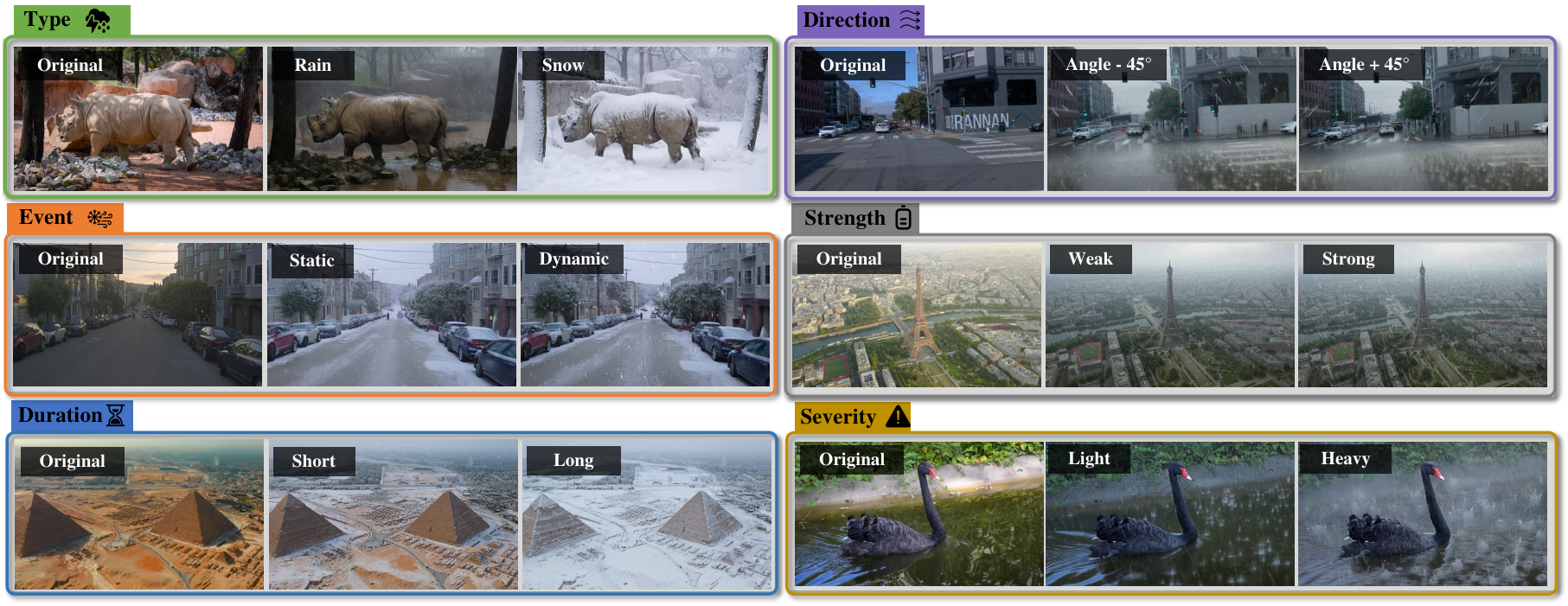}
    \caption{\textbf{Weather Synthesis Examples.} Given an original input video, our method can synthesize diverse weather effects, including the \textbf{\emph{type}} of appearance (snowy or rainy), the \textbf{\emph{event}} condition (static or dynamic), the \textbf{\emph{duration}} (how long the weather lasts), the \textbf{\emph{direction}} and \textbf{\emph{strength}} of wind influence on particles, and the overall \textbf{\emph{severity}}. This enables precise and interpretable control over atmospheric appearance and particle dynamics, yielding diverse and realistic weather effects.}
    \label{fig:fig1}
\end{figure}

\begin{abstract}
Weather synthesis aims to add weather effects to input videos while preserving scene identity, structure, and motion. The key limitation of existing methods is the lack of diversity in weather appearance and effective control over weather dynamics (e.g., temporal evolution and particle motion). Most approaches rely on text prompts, which are inherently underspecified and often fail to produce detailed weather characteristics. Additionally, general-purpose video editors optimized for clean and aesthetic outputs tend to suppress heavy weather phenomena, making dense particle effects difficult to generate.
To address these, we propose a \textit{Semantic-Aware, Physics-Informed, and Geometry-Grounded} framework that steers an off-the-shelf video editor to synthesize diverse global appearances and detailed particle dynamics. We factorize the synthesis into three conditional signals, so that each provides a distinct and stable source of guidance: \textit{\textbf{semantics}} specifies what the weather should look like, \textit{\textbf{dynamics}} governs how it evolves over time, and \textit{\textbf{geometry}} determines where it should appear in the scene. Specifically, we introduce (1) semantic-aware appearance anchoring to establish the target appearance from scene semantics and user input; (2) physics-informed dynamic simulation to generate particle effects by simulating a Gaussian-represented particle field under gravity, wind, and turbulence; and (3) geometry-grounded video synthesis to align the simulated particles with target scene geometry and synthesize the final video. Experiments demonstrate that our method produces diverse, physically and visually realistic weather effects. Furthermore, we show that our synthesized data significantly improves the robustness of autonomous driving semantic segmentation under adverse weather conditions. Project page: \url{https://jumponthemoon.github.io/w-crafter/}.

\end{abstract}

\section{Introduction}
\label{sec:intro}
What would the Colosseum look like blanketed in heavy snowfall, or a desert pyramid drenched by a torrential downpour? Such conditions are rarely captured in the real world, yet can be envisioned through weather synthesis, where weather effects are synthesized and added to existing videos. This capability is valuable for creative applications, where one wishes to control specific weather conditions on real footage~\cite{dai2025rainygs,xie2025drivebench}. Moreover, it can serve as a data augmentation tool for training and evaluating perception systems, enhancing the robustness of autonomous driving and robotic systems under diverse weather conditions~\cite{alhaija2025cosmos,kong2024robodepth}.

Existing methods of weather video synthesis mainly follow two directions. Reconstruction-based pipelines\cite{qian2025weatheredit,li2023climatenerf} simulate weather effects in a 3D scene using NeRF~\cite{mildenhall2021nerf} or 3DGS~\cite{kerbl20233d}, then re-render the scene under synthesized conditions. While they offer interpretable control, the handcrafted effects lack diversity and are sensitive to reconstruction errors, resulting in unrealistic results.
Data-driven approach \cite{lin2025controllable} trains diffusion models on curated weather datasets to directly synthesize effects conditioned on the input video. Although these methods enable style control, the resulting appearance is often biased toward the training distribution, which limits diversity. Furthermore, synthesizing novel weather effects typically requires extensive additional data and costly re-training.

\begin{figure}[H]
    \centering
   \includegraphics[width=0.98\linewidth]{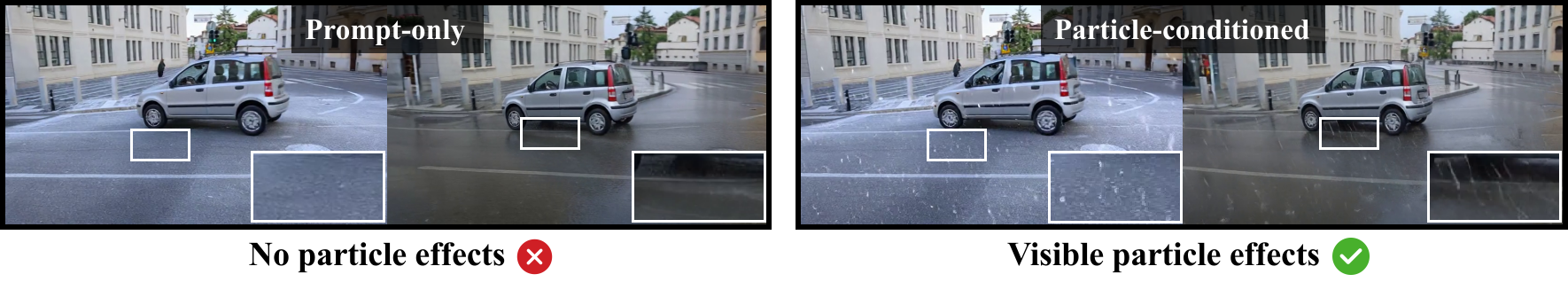}
\vspace{-3mm}
    \caption{\textbf{Comparison between text prompt-only and particle-conditioned editing.} Text prompts alone often yield mild precipitation with weak particle details. In contrast, particle-conditioned editing enables dense snowflakes and rain streaks synthesis, indicating it effectively activates latent weather priors in pretrained video editors.}
\label{fig:fig2}
\vspace{-5mm}
\end{figure}

More recently, conditional video editors incorporate inputs such as depth~\cite{jiang2025vace, HaCohen2024LTXVideo}, optical flow~\cite{chen2025contextflow,liu2025stablev2v} and pose~\cite{ma2024follow,gan2025humandit} together with text prompts, allowing flexible controllability for video-to-video translation \cite{survey_3d_4d_world_models,worldlens}. However, these models are typically trained on clean, high-quality videos, where heavy-weather degradations are under-represented. As a result, prompting for ``heavy snow'' or ``intense rain'' often yields only mild effects, failing to produce dense particles and realistic dynamics. \textit{Interestingly}, we observe that modern video editor models still encode non-trivial priors about particle dynamics (\eg, falling snowflakes or rain streaks). In particular, injecting particle visual primitives and trajectories into structured conditioning (\eg, RGB or depth) can activate particle-aware generation: once a particle-augmented input is provided, the pretrained model can synthesize raindrops and snowflakes without finetuning, as illustrated in Fig.~\ref{fig:fig2}. This suggests that the challenge is not whether such models can generate weather particles, but how to \emph{activate} this behavior in an effective manner. 

Our key insight is that a well-designed conditioning space is essential: structured signals that jointly encode appearance and dynamics provide explicit guidance beyond text-only prompts. Motivated by this, we propose a \emph{Semantic-Aware, Physics-Informed, and Geometry-Grounded} framework. Specifically, we decompose weather synthesis into three complementary conditioning signals: semantics, dynamics, and geometry, instantiated as semantic-aware appearance anchoring, physics-informed dynamic simulation, and geometry-grounded video synthesis. For \emph{\textbf{semantics}}, we leverage a Multimodal Large Language Model (MLLM) \cite{achiam2023gpt} to construct appearance anchors that define the target weather style based on user intent. For \emph{\textbf{dynamics}}, we simulate a particle field represented by Gaussian primitives under physical priors, governing particle motion and temporal behavior. For \emph{\textbf{geometry}}, we fuse scene and weather particles with geometry to form particle-augmented depths as conditioning for video synthesis. By combining these three conditioning inputs, we steer an off-the-shelf video diffusion model to synthesize diverse and dynamic weather effects with high controllability, as shown in ~\cref{fig:fig1}. Our contributions are fourfold:
\begin{itemize}
    \item A \textbf{\emph{novel and controllable weather synthesis framework}} that enables diverse global appearances and fine-grained particle dynamics by conditioning on structured signals of semantics, dynamics, and geometry.
    \item A \textbf{\emph{semantic-aware editing strategy}} that binds intended weather effects to scene semantics, improving controllability and reliability, and yielding diverse and realistic synthesis results.
    \item A \textbf{\emph{physics-informed particle representation}} that effectively activates latent weather priors in pretrained diffusion models, producing temporally coherent and physically plausible dynamic effects.
    \item A \textbf{\emph{geometry-grounded mechanism}} that couples particle dynamics with 3D scene geometry, ensuring temporally consistent trajectories and spatially accurate placement across video frames.
\end{itemize}
\section{Related Work}
\label{sec:related_work}

\textbf{Weather Effect Synthesis.} This task has been extensively explored in both graphics and vision \cite{robodrive_challenge_2024,kong2023robo3d,xie2025benchmarking}. Conventional approaches model precipitation and atmospherics through particle-based simulation~\cite{stomakhin2013material,gissler2020implicit,tremblay2021rain} and physical scattering models~\cite{nayar1999vision,hahner2019semantic,hao2024is}. Learning-based methods~\cite{qian2025weatherdg, schmidt2021climategan,li2021weather,parmar2024one,hao2025safemap,kong2025multi} further perform weather translation or augmentation from data, enabling flexible appearance changes without explicit simulation. With the emergence of neural rendering, several works~\cite{qian2025weatheredit, li2023climatenerf,xie2025climategs,fiebelman2025let,sang2025weather,wu2026weathercity} inject weather into implicit or explicit 3D neural representations and re-render scenes under simulated conditions. These pipelines offer interpretable controls, but often require costly per-scene optimization and are sensitive to reconstruction errors. More recently, WeatherWeaver~\cite{lin2025controllable} trains a weather-specific video diffusion model using synthetic datasets curated by graphics engines\cite{engine2018unreal}, but the resulting appearance is tied to the simulator, and transferring to new domains or styles typically requires additional data.

Unlike reconstruction-based pipelines and simulator-curated editors, we neither require costly per-scene optimization nor rely on fixed training distributions. Instead, we introduce a semantic-aware anchoring strategy to define scene-specific atmospheric styles and leverage physics-simulated particles to govern realistic weather dynamics. By grounding these elements with 3D geometric cues, we enable a general video diffusion model to render diverse and consistent weather effects in a zero-shot manner.

\noindent\textbf{Conditional Video Editing.} Conditional video editing performs video-to-video translation under user control while preserving identity and motion~\cite{feng2024ccedit,ku2024anyv2v,jones2026tuning,gao2026pisco}. Existing methods achieve control by conditioning generative models on structured signals such as depth\cite{xing2024make,guo2024sparsectrl,chen2023control}, optical flow\cite{liu2025stablev2v,cong2023flatten}, camera trajectories\cite{bahmani2024vd3d,kuang2024collaborative,bai2025recammaster}, or pose\cite{gan2025humandit,pang2025dreamdance,ma2024follow}. Recent works further support multi-condition inputs for richer, more compositional control~\cite{lu2025infinicube,jiang2025vace,alhaija2025cosmos,ali2025world}. Compared with prompt-only editing \cite{kong2024hunyuanvideo,yang2024cogvideox,wang2025videodirector,chen2025contextflow,chow2025editmgt}, such structured conditions better constrain how visual effects change over time, improving alignment with scene structure. 

The primary limitation of these methods is the inability to achieve \textbf{fine-grained semantic and temporal control over weather}. Most general video editors are targeting clean, high-quality outputs; heavy weather degradations are under-represented and hard to elicit reliably, resulting in limited weather appearance, weak particle details, and unstable temporal dynamics during editing. This motivates our structured conditioning interface, which provides stable guidance by injecting appearance semantics, particle dynamics, and 3D geometric cues into a video editor for diverse and consistent weather synthesis.



\section{Methodology}
\label{sec:method}

\begin{figure}[t]
    \centering
    \includegraphics[width=\linewidth]{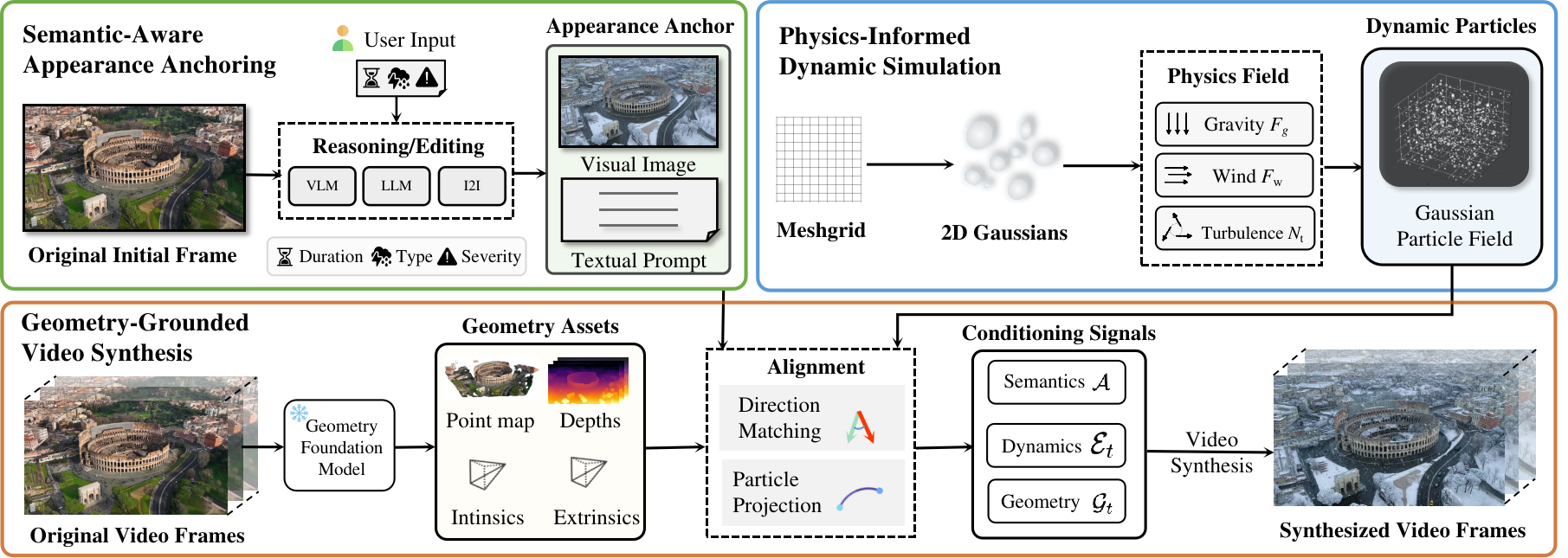}
    \caption{\textbf{Overview of our framework.} Given an input video, we construct structured conditioning signals from three aspects: \emph{semantics} (semantic-aware appearance anchoring), \emph{dynamics} (physics-informed dynamic simulation), and \emph{geometry} (geometry-grounded video synthesis). These signals are combined to guide a video diffusion model to synthesize realistic weather effects.}
    \label{fig:fig3}
\end{figure}

Our framework consists of three stages (Fig.~\ref{fig:fig3}):
\textbf{(i) Semantic-aware appearance anchoring} constructs an appearance anchor composed of an edited initial-frame reference and a scene-grounded text prompt. Specifically, our MLLM-based pipeline employs a Vision-Language Model (VLM) to parse the scene, a Large Language Model (LLM) to refine the instructions, and an image-to-image (I2I) model to synthesize the initial frame.
\textbf{(ii) Physics-informed particle simulation} simulates a weather particle field under physically motivated forces (\eg, gravity and wind), producing temporally coherent particle cues that encode dynamics.
\textbf{(iii) Geometry-grounded video synthesis} extracts per-frame geometry from the input video and uses this information to align the outputs from previous stages.
This yields the final conditioning signals: semantics $\mathcal{A}$ with edited initial-frame and text prompt, dynamics $\{\mathcal{E}_t\}$ with particle-augmented depth maps, and geometry $\{\mathcal{G}_t\}$ with original scene depths, which jointly condition the video diffusion editor to produce the final video $\tilde{V}$.

\subsection{Semantic-Aware Appearance Anchoring}
\label{sec:appearance}
\begin{figure}[t]
    \centering
    \includegraphics[width=\linewidth]{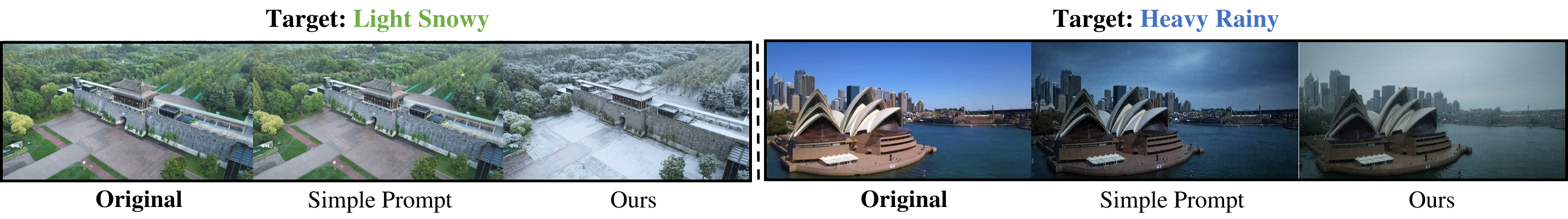}
    \caption{\textbf{Comparisons of editing results between the simple prompt and our prompt.} With the simple prompt, the editing model struggles to produce light snow with clearly visible effects and heavy rain with accurate, scene-consistent lighting. In contrast, our generated prompt incorporates richer scene semantics and finer-grained details, leading to outputs that more reliably match the intended conditions.}
    \label{fig:fig4}
\end{figure}

\begin{figure}[t]
    \centering
    \includegraphics[width=0.95\linewidth]{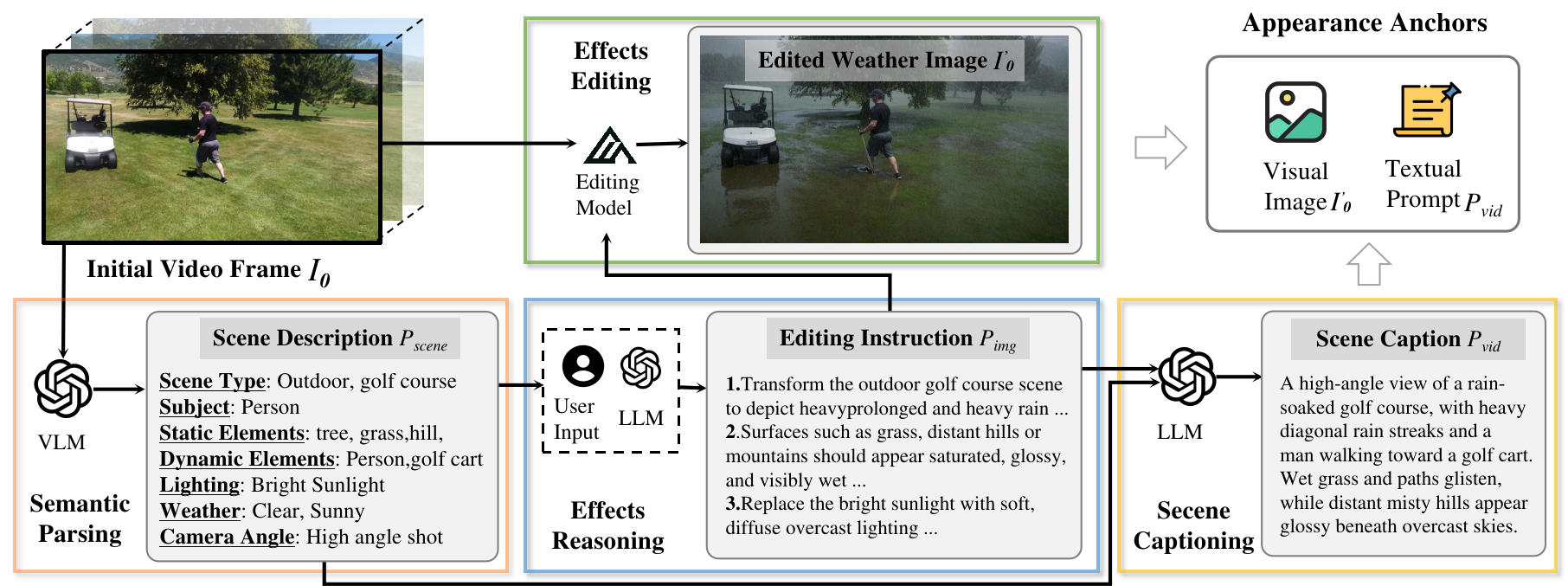}
   \caption{\textbf{Semantic-aware appearance anchoring.} Given the initial video frame, a VLM parses scene semantics and an LLM reasons about weather-specific effects conditioned on user intent. The resulting instruction is used to edit the initial frame to anchor target appearance, while also being combined with scene semantics to provide a refined text description for subsequent video generation.}
   \label{fig:fig5}
\end{figure}

To guide weather video synthesis toward a desired target appearance, we first edit the video’s initial frame using an image editing model. However, a naive prompt such as \emph{``make it snowy''} is inherently underspecified, offering limited control over key factors such as severity and lighting tone. Even with coarse modifiers (\eg, “heavy” or “light”), prompts remain ambiguous and often lead to limited editing effects (Fig.~\ref{fig:fig4}). We therefore develop a semantic-aware editing strategy comprising: semantic parsing, effects reasoning, and scene captioning. Our core idea is to craft detailed editing prompts by explicitly binding global appearance changes to scene semantics based on user intent, enabling fine-grained control and achieving desirable editing outcomes, as shown in Fig.~\ref{fig:fig5}.

\noindent\textbf{Semantic Parsing.}
Given the initial frame $I_0$, we use a VLM to translate the visual scene into interpretable attributes and produce a structured description $P_\text{scene}$ 
via $P_\text{scene} \leftarrow \mathrm{VLM}(I_0)$.
This structured representation grounds subsequent edits in scene semantics, enabling precise and localized appearance changes that correctly reflect the intended weather. 

\noindent\textbf{Effects Reasoning.} Conditioned on the user intent $\mathcal{U}$ and the scene description $P_\text{scene}$, we employ an LLM to generate an image editing instruction $P_{\mathrm{img}}$.
In particular, $\mathcal{U}$ specifies target weather type, duration, and severity, while $P_\text{scene}$ provides scene context to ensure the instruction is semantically plausible, yielding $P_{\mathrm{img}} \leftarrow \mathrm{LLM} (P_{\mathrm{scene}},\mathcal{U})$. We then generate an edited image ${I'}_0$ using an I2I model $\mathcal{M}_{\mathrm{img}}$, which establishes scene-level weather appearance (\eg, wet surface) together with a consistent global lighting tone (\eg, overcast illumination):
\begin{equation}
{I'}_0 \leftarrow \mathcal{M}_{\mathrm{img}}(I_0, P_{\mathrm{img}}).
\label{eq:appearance_anchor}
\end{equation}

\noindent\textbf{Scene Captioning.}
To provide a compact and consistent target for subsequent video synthesis, we further produce a video editing prompt $P_{\mathrm{vid}}$ with a scene caption that summarizes the expected edited scene under the target weather, capturing global atmosphere and key dynamic cues:
\begin{equation}
P_{\mathrm{vid}} \leftarrow \mathrm{LLM}(P_{\mathrm{scene}}, P_{\mathrm{img}}).
\label{eq:event_caption}
\end{equation}



At the subsequent video synthesis stage, we integrate explicit weather cues by projecting the particle effects into ${I}'_0$, yielding $\tilde{I}_0$ and feed it together with $P_{\mathrm{vid}}$ as the final appearance anchor $\mathcal{A}$ to the video editing model. The first-frame $\tilde{I}_0$ serves as a strong appearance reference that initializes the scene structure and global style, while $P_{\mathrm{vid}}$ supplies a high-level description that encourages temporally consistent semantics across frames.

\begin{figure}[t]
    \centering
    \includegraphics[width=\linewidth]{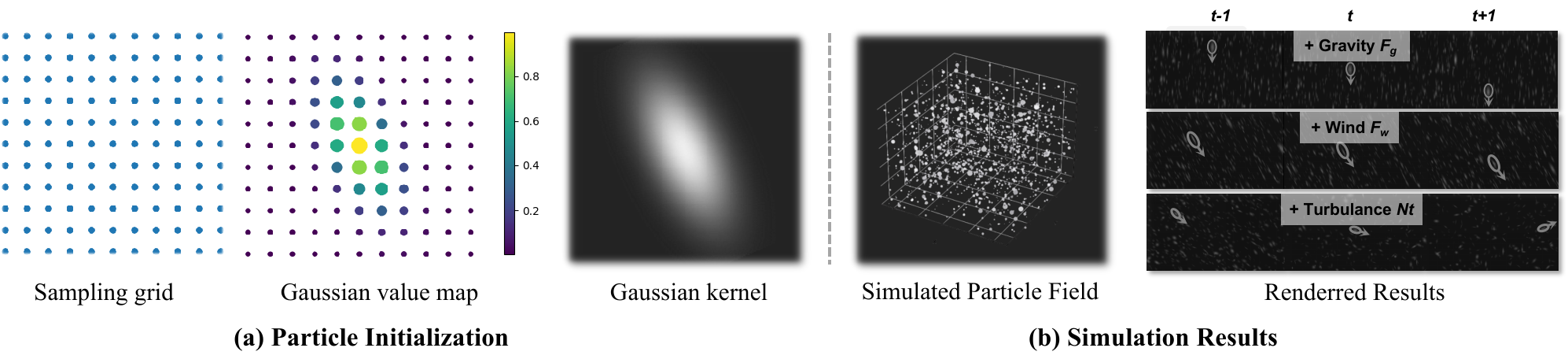}
\caption{\textbf{Physics-informed dynamic simulation.}
(a) Weather particles are initialized as grid-sampled anisotropic Gaussians.
(b) The particle field is evolved under gravity, wind, and turbulence, producing physically plausible motion that serves as explicit motion cues for video synthesis.
}
    \label{fig:fig6}
\end{figure}

\subsection{Physics-Informed Dynamic Simulation}
\label{sec:physics2d}

To generate dynamic weather effects, we model weather as a set of temporally persistent particles in a local field. Each particle is represented as an anisotropic Gaussian primitive (Fig.~\ref{fig:fig6}a), and its motion is governed by physics priors such as gravity, wind, and turbulence (Fig.~\ref{fig:fig6}b). This produces a physically plausible and controllable dynamic weather representation that can be geometrically grounded to video frames in later stages.

\noindent\textbf{Particle initialization.}
We define a local sampling grid $\{\mathbf{s}\}$ with $\mathbf{s}=(x,y)\in\mathbb{R}^2$ and represent the appearance of particle $i$ by an anisotropic Gaussian kernel:
\begin{equation}
G_i(\mathbf{s})=\exp\!\left(
-\frac{x^2}{2(\sigma_{x,i}^2)\,\gamma\,}
-\frac{y^2}{2(\sigma_{y,i}^2)\,\gamma\,}
\right),
\label{eq:kernel_2d}
\end{equation}
where $(\sigma_{x,i},\sigma_{y,i})$ parameterize per-particle anisotropy and $\gamma$ controls the depth-dependent scale that is derived from the particle depth during projection.
The resulting stamp serves as a compact primitive for fine-grained particle appearance (\eg, rounded snowflakes or elongated raindrops) in the subsequent simulation.

\noindent \textbf{Physics-informed simulation.}
To simulate particle dynamics, we augment each particle's state with a depth coordinate, i.e., $\mathbf{x}_i=(X_i,Y_i,Z_i)\in\mathbb{R}^3$, forming a particle field.
We evolve each particle with a force-based dynamics that serves as an explicit temporal prior:
\begin{equation}
\dot{\mathbf{x}}_i(t)=\mathbf{v}_i(t),\qquad
m\,\dot{\mathbf{v}}_i(t)=\mathbf{F}_{g}+\mathbf{F}_{w}+\mathbf{N}_{t}(\mathbf{x}_i(t),t),
\label{eq:newton_dynamics}
\end{equation}
where $\dot{(\cdot)}$ denotes the time derivative, $\mathbf{v}_i(t)$ corresponds to the velocity of particle $i$ at time $t$, and $m$ represents its mass. Gravity is modeled as $\mathbf{F}_g = m g\,\mathbf{d}_g$, with magnitude $g$ and direction $\mathbf{d}_g$, while wind is parameterized as $\mathbf{F}_{w}=\alpha\,\mathbf{d}_w$ with strength $\alpha$ and direction $\mathbf{d}_w$.
To capture stochastic local perturbations, we model turbulence as $\mathbf{N}_{t}(\mathbf{x},t)=\beta\,\mathbf{n}(\mathbf{x},t)$, where $\beta$ controls its strength and $\mathbf{n}$ is a divergence-free field instantiated via curl noise:
\begin{equation}
\mathbf{n}(\mathbf{x},t)=\nabla\times\mathbf{\psi}(\mathbf{x},t), \quad \nabla\!\cdot\!\mathbf{n}(\mathbf{x},t)=0,
\label{eq:curl_force}
\end{equation}
where $\mathbf{\psi}$ is derived from a 3D noise field $o(\cdot)$ (\eg, Perlin noise), such as
\begin{equation}
\mathbf{\psi}(x,y,z)=\big(o(x,y,z),\,o(y,z,x),\,o(z,x,y)\big).\label{eq:vector_potential}
\end{equation}
Finally, we discretize Eq.~\eqref{eq:newton_dynamics} using an explicit Euler integrator:
\begin{equation}
\mathbf{v}_i^{t+1}=\mathbf{v}_i^{t}+\frac{\mathbf{F}_w+\mathbf{F}_g+\mathbf{N}_t(\mathbf{x}_i^t,t)}{m}\Delta t,\quad
\mathbf{x}_i^{t+1}=\mathbf{x}_i^{t}+\mathbf{v}_i^{t+1}\Delta t.
\end{equation}
This formulation yields an expressive dynamics prior: gravity and wind govern global motion, while turbulence introduces locally varying, volume-preserving perturbations for realistic particle behavior.

\subsection{Geometry-Grounded Video Synthesis}
\label{sec:geometry}
The simulated particle field is defined in a canonical coordinate system, where gravity is simply a fixed axis. However, the input video is captured in a scene-specific coordinate system: the true ``down'' direction depends on the 3D orientation of the scene. Additionally, the camera undergoes motion across frames in dynamic scenes.
If we directly inject particles into the image plane with a fixed camera in canonical space, raindrops and snowflakes exhibit incorrect falling directions and drift with the camera, leading to implausible weather effects.

\begin{figure}[t]
  \centering
   \includegraphics[width=1.0\linewidth]{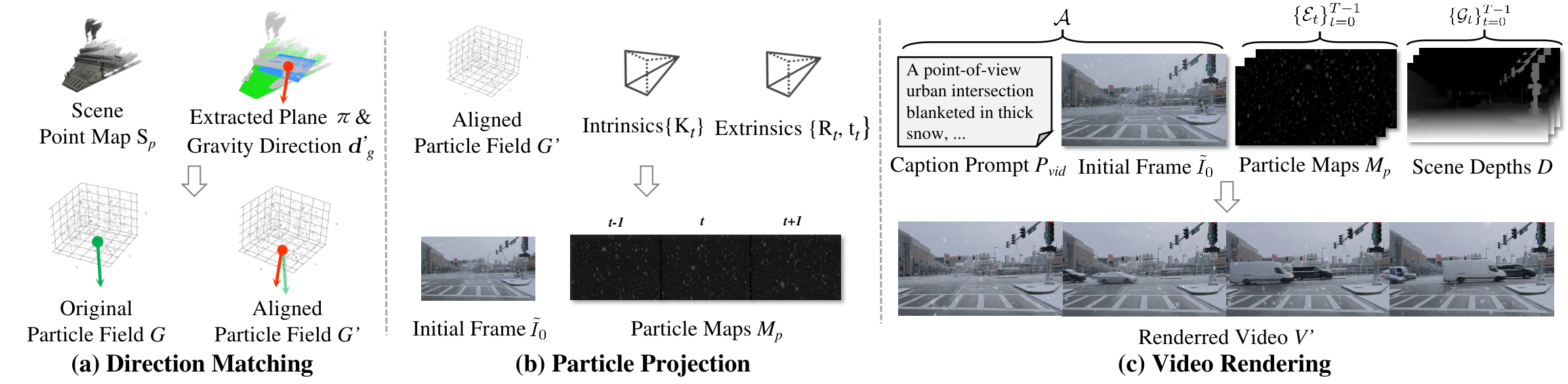}
    \caption{\textbf{Geometry-grounded video synthesis.}
    (a) The simulated particle field is aligned with the scene by estimating gravity direction from 3D geometry.
    (b) The aligned particles are then projected into edited initial frame and each depth frame using camera intrinsics and extrinsics to produce particle maps.
    (c) Finally, we condition video generation on the particle-augmented inputs together with scene geometry and appearance anchor, yielding plausible and consistent weather effects.}
   \label{fig:geometry}
\end{figure}

To address this, we ground both simulation and synthesis in scene geometry (Fig.~\ref{fig:geometry}). In detail, we first employ a geometry foundation model\cite{lin2025depth} to estimate a 3D point map $S_p$, per-frame depth maps $D$, and camera parameters $(K_t,R_t,\mathbf{t}_t)$ from the input video $V$. Then, we align the particle field with the scene geometry by (i) matching its gravity direction with that of the 3D point map, and (ii) projecting the evolving particle field into each frame using the estimated camera parameters. The resulting particle maps, together with scene geometry and appearance anchors, serve as the final conditioning signals for video synthesis.
\noindent\textbf{Direction matching}.
Given a point map $S_p$, we sample 3D points $\mathbf{p}$ and fit the dominant ground plane using point-to-plane distance as the residual. Specifically, we estimate $\boldsymbol{\pi}: \mathbf{c}^{\top}\mathbf{p}+d=0$, and count $\mathbf{p}$ as an inlier if $|\mathbf{c}^{\top}\mathbf{p}+d|<\tau$, where $\mathbf{c}$ is the unit normal and $\tau$ is an inlier threshold. Then, we take the scene gravity direction as $\mathbf{d}_{g}'=-\mathbf{c}$.
Let $\mathbf{d}_{g}$ denote the gravity axis in our canonical simulation. 
We compute the minimal rotation $R_g$ satisfying $R_g \mathbf{d}_{g}=\mathbf{d}_{g}'$,
and apply it to particle positions and velocities to obtain the aligned field $G'$.

\noindent\textbf{Particle projection.}
For each frame $t$, we use the estimated camera intrinsics and extrinsics $(K_t,R_t,\mathbf{t}_t)$ to project particle $i$ with its 3D position $\mathbf{x}_i^t$ into the image plane:
\begin{equation}
\mathbf{u}_{i,t}=\pi\!\left(K_t,R_t,\mathbf{t}_t,\mathbf{x}_i^t\right), \qquad
z_{i,t}=\big(R_t\mathbf{x}_i^t+\mathbf{t}_t\big)_z ,
\label{eq:particle_proj}
\end{equation}
where $\mathbf{u}_{i,t}\in\mathbb{R}^2$ is the pixel coordinate and $z_{i,t}$ is the camera-space depth for scale control.
To couple appearance with dynamics (\eg, raindrops elongate along the motion direction), we orient each particle stamp by the projected velocity direction:
\begin{equation}
\theta_{i,t}=\mathrm{atan2}\!\left((\mathbf{v}_i^t)^y,\ (\mathbf{v}_i^t)^x\right),
\label{eq:stamp_orient}
\end{equation}
yielding particle maps $M_p$ that serve as dynamic cues $\mathcal{E}_t$. We also composite the particles onto the edited initial frame as $\tilde{I}_0$ to update appearance anchor $\mathcal{A}$.

\noindent\textbf{Video Synthesis.}
Finally, we steer a pretrained conditional video editor $\mathcal{M}_{vid}$ with the appearance anchor $\mathcal{A}$ , the particle dynamics $\{\mathcal{E}_t\}$, and the geometric cues $\{\mathcal{G}_t\}$ to synthesize final video:
\begin{equation}
\tilde{V}=\mathcal{M}_{\mathrm{vid}}\big(V;\ \mathcal{A},\{\mathcal{E}_t\}_{t=0}^{T-1},\{\mathcal{G}_t\}_{t=0}^{T-1}\big).
\label{eq:video_render}
\end{equation}
The resulting tri-prior conditioning yields disentangled control over global atmosphere, particle dynamics, and geometry-consistent placement. This bridges explicit physical simulation and implicit generative modeling, enabling realistic weather synthesis with flexible control. 
\section{Experiments}
\begin{figure}[t]
  \centering
   \includegraphics[width=\linewidth]{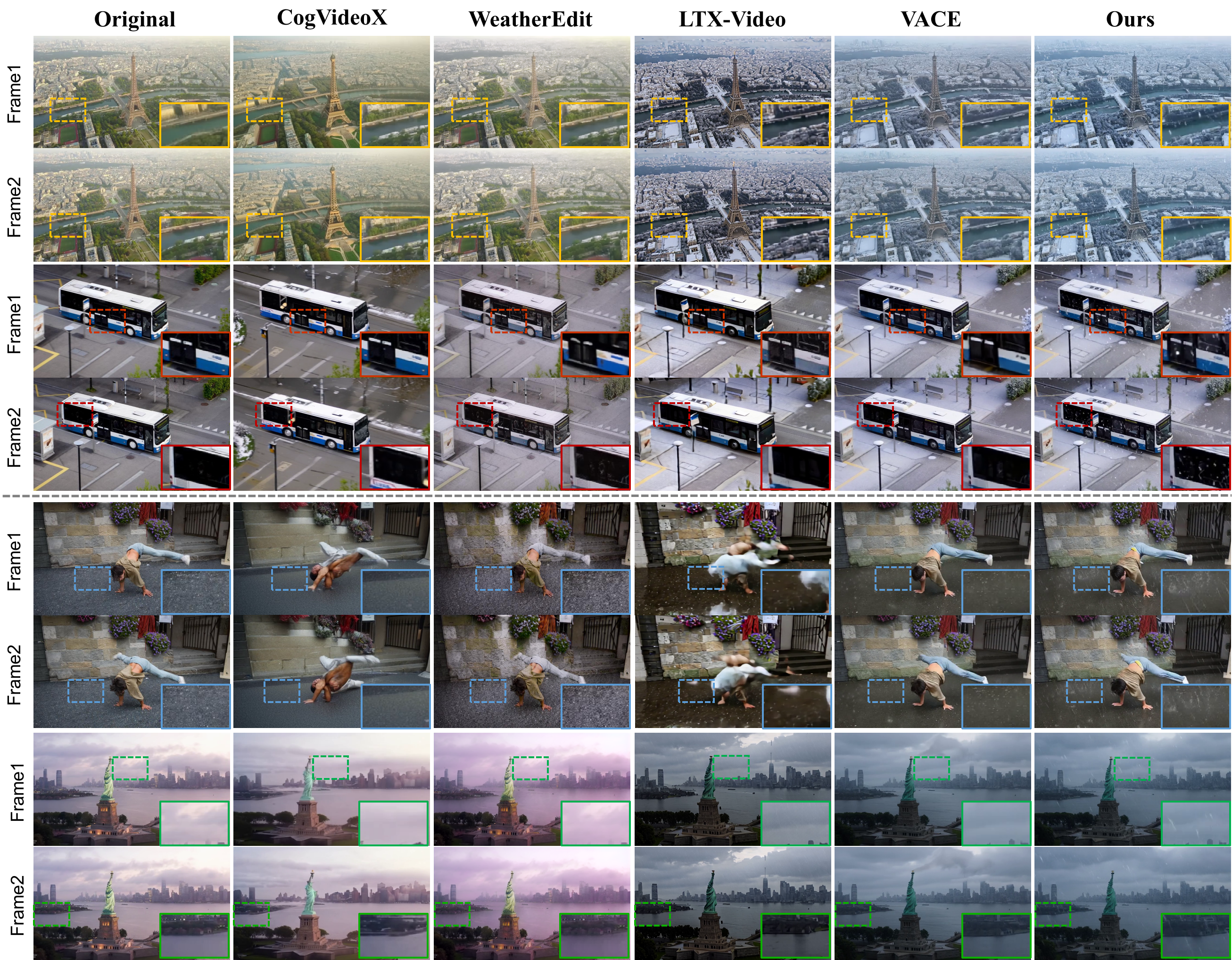}
    \caption{\textbf{Qualitative comparison under different weather conditions.} Top: snowy scenes. Bottom: rainy scenes. Particle details are highlighted in colored boxes.}
    \label{fig:sota}
   \label{fig:weather_effect}
\end{figure}
\subsection{Implementation Details}
Our implementation follows the three-stage design in \cref{sec:method}: appearance anchoring, particle simulation, and geometry grounding. For appearance anchoring, we use GPT-4.1~\cite{achiam2023gpt} to parse the initial frame into a structured scene description $P_{\text{scene}}$, and to reason about the target weather effects, yielding an image-editing instruction and a compact video prompt; FLUX.1-Kontext Pro~\cite{labs2025flux} then edits the initial frame to produce the appearance anchor. For geometry grounding, we adopt Depth Anything V3~\cite{lin2025depth} to obtain geometry information and drive the particle simulation. We use Wan2.1-VACE~\cite{jiang2025vace} as the conditional video diffusion model with its default inference settings. The preprocessing steps take $\sim$3 minutes per video, while the overall runtime is dominated by video synthesis at $\sim$38 minutes. All experiments are conducted on a single NVIDIA RTX A6000 GPU.

\subsection{Dataset}
To evaluate generalizability and robustness, we curate an evaluation dataset consisting of 58 videos from DAVIS\cite{perazzi2016benchmark}, DL3DV-10k\cite{ling2024dl3dv}, PandaSet\cite{xiao2021pandaset}, and nuScenes\cite{caesar2020nuscenes}. It spans a wide range of real-world scenarios captured from aerial and ego-centric perspectives, encompassing both static and dynamic camera setups with diverse motion trajectories. This ensures diversity in scene types, geometric complexity, and illumination, all of which directly affect atmospheric appearance and particle visibility. All sequences are temporally trimmed and processed with a unified preprocessing protocol for fair evaluation. Additionally, we use ACDC~\cite{sakaridis2021acdc} and MUSES~\cite{brodermann2024muses} for downstream task evaluation, covering diverse weather conditions in autonomous driving scenarios.


\subsection{Evaluation Metrics}
For each input video, we evaluate synthesized rain and snow effects using metrics that measure semantic preservation, prompt alignment, and perceptual quality. For semantic preservation, we measure CLIP similarity\cite{radford2021learning} (CLIP-S) between edited and input videos. For prompt alignment, we compute CLIP directional alignment\cite{gal2022stylegan} (CLIP-D) between edited frames and the editing prompt. We further employ a VLM model GPT-4V\cite{achiam2023gpt} for evaluating high-level semantic alignment and overall perceived quality. In addition, we conduct a human evaluation to assess perceptual realism and collect user preferences for selecting the best result. Finally, we report mIoU for semantic segmentation evaluation. 



\begin{table}[t]
\centering
\caption{\textbf{Quantitative evaluation under different weather conditions.} 
We conduct CLIP-based and VLM-based evaluations for snow and rain settings. The overall score is averaged across conditions.}
\setlength{\tabcolsep}{3.0pt}
\renewcommand{\arraystretch}{1.15}
\small
\resizebox{\columnwidth}{!}{
\begin{tabular}{lccccccccc}
\toprule
\multirow{2}{*}{\textbf{Method}} 

& \multicolumn{3}{c}{\textbf{Snow}} 
& \multicolumn{3}{c}{\textbf{Rain}} 
& \multicolumn{3}{c}{\textbf{Overall}} \\
\cmidrule(lr){2-4} \cmidrule(lr){5-7} \cmidrule(lr){8-10}

& CLIP-D $\uparrow$ & CLIP-S $\uparrow$ & VLM $\uparrow$
& CLIP-D $\uparrow$ & CLIP-S $\uparrow$ & VLM $\uparrow$
& CLIP-D $\uparrow$ & CLIP-S $\uparrow$ & VLM $\uparrow$ \\
\midrule
CogVideoX\cite{yang2024cogvideox}
& 0.08 & 0.82 & 0.56 
& 0.06 & 0.82 & 0.52 
& 0.07 & 0.82 & 0.54 \\

WeatherEdit\cite{qian2025weatheredit} 
& 0.14 & \textbf{0.90} & 0.45 
& 0.12 & \textbf{0.92} & 0.36 
& 0.13 & \textbf{0.91} & 0.41 \\

LTX-Video\cite{HaCohen2024LTXVideo}
& 0.15 & 0.80 & 0.68 
& 0.12 & 0.83 & 0.67 
& 0.13 & 0.82 & 0.67 \\

VACE\cite{jiang2025vace} 
& 0.18 & 0.80 & 0.76 
& 0.14 & 0.83 & 0.74 
& 0.16 & 0.82 & 0.75 \\

\midrule
\rowcolor{gray!5}
\textbf{Ours} 
&\textbf{ 0.19} & 0.79 & \textbf{0.84} 
&\textbf{ 0.15}& 0.82 & \textbf{0.81} 
& \textbf{0.17}& 0.81 & \textbf{0.83} \\
\bottomrule

\end{tabular}
}
\label{tab:quantitive}
\end{table}

\subsection{Results}
We compare our method against two groups of representative video editing methods.
The first stream consists of prompt-only methods, such as CogVideoX\cite{yang2024cogvideox} and WeatherEdit\cite{qian2025weatheredit}, which take an input RGB video along with a text prompt to generate the edited sequence.
The second stream includes conditional editors, such as LTX-Video~\cite{HaCohen2024LTXVideo} and VACE~\cite{jiang2025vace}, which incorporate inputs including reference images and depth maps, to guide the editing process.
For a fair comparison on fine-grained particle effects, we provide all conditional baselines with the same text prompt as our method and the same reference image but with particle conditioning removed.

\noindent\textbf{Qualitative comparison.}
For snowy scenes, prompt-only methods such as CogVideoX\cite{yang2024cogvideox} fail to produce convincing snow effects, while WeatherEdit\cite{qian2025weatheredit} only yields mild whitening of the input, as illustrated in Fig.~\ref{fig:sota}. Although conditioning-based editors such as LTX-Video\cite{HaCohen2024LTXVideo} and VACE\cite{jiang2025vace} better match the global snowy tone, they rarely synthesize dense particle effects with falling snowflakes.
For rainy weather, CogVideoX\cite{yang2024cogvideox} and WeatherEdit\cite{qian2025weatheredit} darken the scene and increase surface reflectance (\eg, the dancing sequence), yet their edits are inconsistent across scenes (\eg, limited changes on the statue scene).
Although LTX-Video\cite{HaCohen2024LTXVideo} and VACE\cite{jiang2025vace} produce global tone change, they fail to synthesize visible rain streaks in a stable manner. 

In general, prompt-only editing tends to underexpress heavy weather degradations, while conditional editors often prioritize background appearance and suppress fine-grained particles. In contrast, our method can effectively edit both the general atmosphere and detail particles, producing realistic weather synthesis results across diverse conditions.

\noindent \textbf{Quantitative comparison.}
As shown in Tab.~\ref{tab:quantitive}, our method achieves the highest CLIP-D score across all settings, indicating the closest alignment between the synthesized weather effects and the desired target conditions. 
Moreover, the top ratings from the VLM-based evaluator suggest superior overall perceptual quality, with more convincing and visually coherent weather phenomena. Notably, WeatherEdit\cite{qian2025weatheredit} attains the highest CLIP-S score, reflecting strong preservation of the original scene. However, this largely stems from its conservative editing behavior, which introduces only subtle weather effects. 
VACE\cite{jiang2025vace} and LTX-Video\cite{HaCohen2024LTXVideo} exhibit a similar trend: they maintain structure consistency, yet produce weaker and less realistic weather effects, as evidenced by lower CLIP-D scores compared to ours. 
In contrast, our method introduces substantial atmospheric changes with dense particle effects while largely preserving scene identity, achieving a balance between weather expressiveness and scene fidelity.
\begin{table}[t]
\centering
\begin{minipage}[t]{0.56\linewidth}
\centering
\caption{\textbf{Realism rating.} The rating is based on a 5-point Likert scale. Our method achieves the highest scores.}
\resizebox{\linewidth}{!}{%
\begin{tabular}{lcc}
\toprule
\textbf{Method} & \textbf{Photo-realism}$\uparrow$ & \textbf{Physical-realism}$\uparrow$ \\
\midrule
CogVideoX\cite{yang2024cogvideox}         & 2.51          & 2.66 \\
WeatherEdit\cite{qian2025weatheredit}     & 2.50          & 2.54 \\
LTX-Video\cite{HaCohen2024LTXVideo}       & 2.87          & 2.72 \\
VACE\cite{jiang2025vace}                  & 3.58          & 3.52 \\
\midrule
\textbf{Ours}                              & \textbf{4.16} & \textbf{4.12} \\
\bottomrule
\end{tabular}}
\label{tab:realism}
\end{minipage}
\hfill
\begin{minipage}[t]{0.40\linewidth}
\centering
\caption{\textbf{User preference.} Percentage of selections as best under rain and snow.}
\resizebox{\linewidth}{!}{%
\begin{tabular}{lcc}
\toprule
\textbf{Method} & \textbf{Rain(\%)} & \textbf{Snow(\%)} \\
\midrule
CogVideoX\cite{yang2024cogvideox}         & 8.7  & 2.2  \\
WeatherEdit\cite{qian2025weatheredit}    & 15.2 & 10.9 \\
LTX-Video\cite{HaCohen2024LTXVideo}      & 6.5  & 4.3  \\
VACE\cite{jiang2025vace}                 & 28.3 & 26.1 \\
\midrule
\textbf{Ours}                             & \textbf{41.3} & \textbf{56.5} \\
\bottomrule
\end{tabular}}
\label{tab:user}
\end{minipage}
\end{table}

\noindent \textbf{Human Evaluation.} 
To assess both photorealism and physical realism of the generated videos, we follow prior work\cite{liu2024physgen} to conduct human evaluation. Photorealism evaluates perceptual quality and alignment with the target condition, while physical realism measures whether the weather dynamics (\eg, particle motion and scene interactions) are physically plausible. We recruit 46 participants and ask them to rate each video on a 5-point Likert scale with 1 indicating strongly disagree and 5 denoting strongly agree for the statements “the video is physically realistic” and “the video is photorealistic”. To reduce ordering bias, the results are presented in random order. In Tab.~\ref{tab:realism}, our method receives the highest average ratings on both criteria, indicating strong physical plausibility and visual quality. We additionally collect the user preference results in which participants select the most preferred edited result among competing methods. As shown in Tab.~\ref{tab:user}, our method is selected most frequently under both weather types, receiving 41.3\% of votes in rain and 56.5\% in snow. This suggests a clear advantage in overall perceived quality, especially for more challenging snow cases where fine-grained particles and realistic dynamics are critical.

\begin{figure}[t!]
    \centering
    \includegraphics[width=\linewidth]{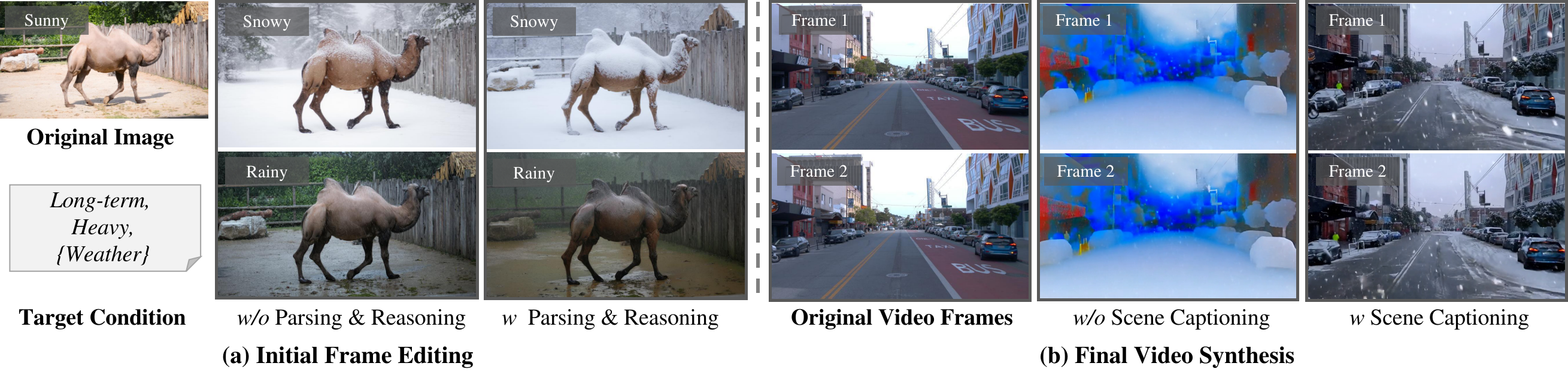}
    \caption{\textbf{Ablation on appearance anchoring.} (a) Without semantic parsing and effect reasoning to generate a detailed prompt, results either change background content (snowy) or produce limited effects on the subject (rainy). With our prompt, the desired effects are achieved while retaining scene structure. (b) Without detailed prompts derived from scene captioning, results often exhibit hallucinations. With captioning, the model preserves scene identity and faithfully reflects the intended weather edits.} 
   \vspace{-3mm}

   \label{fig:ablation_semantic}
\end{figure}
\begin{figure}[t!]
    \centering
    \includegraphics[width=\linewidth]{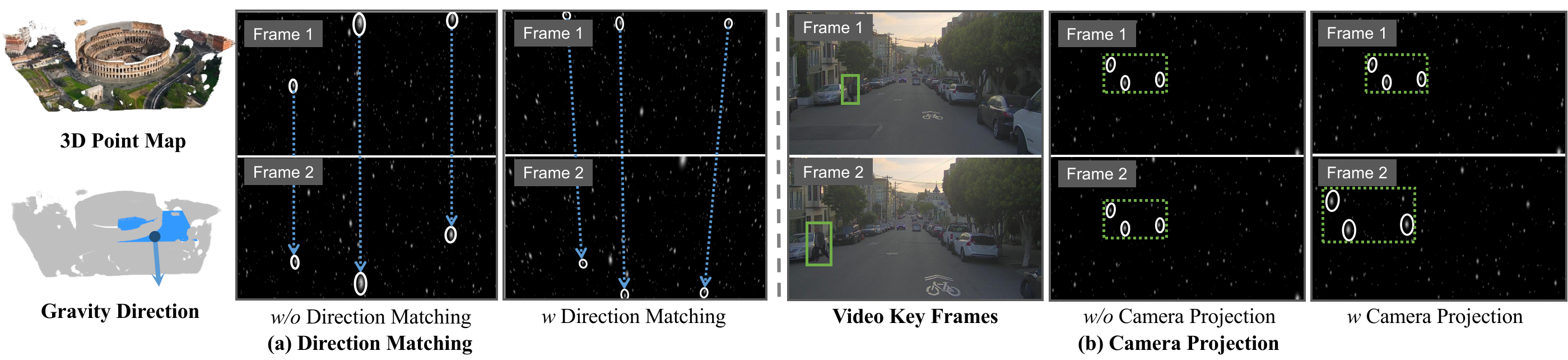}
    \caption{\textbf{Ablation on geometry grounding.} Representative particles are highlighted in white. (Left) The blue solid arrow denotes the true gravity direction in 3D, and the dotted arrows show falling trends in 2D. (Right) The green solid boxes indicate the reference object scale, while the dashed boxes represent the particle-group scale.}
   \label{fig:ablation_geometry}
   \vspace{-3mm}

\end{figure}
\subsection{Ablation Study}
We validate our key designs in (i) appearance anchoring (semantic parsing, effect reasoning, and scene captioning) and (ii) geometry grounding (gravity direction matching and particle camera projection), indicating that each component is critical to achieving diverse, dynamic, and physically plausible weather synthesis.

\noindent\textbf{Appearance anchoring.} In Fig.~\ref{fig:ablation_semantic}, for initial frame editing, we compare a naive prompt (\eg, ``make it long-term, heavy \{weather\}'') with a detailed prompt generated by semantic parsing and effects reasoning. Under snow, the naive prompt exhibits hallucinations and alters the background. Under rain, it fails to induce heavy precipitation, leaving the camel visually dry. In contrast, our prompt yields more faithful atmospheric cues including low-lighting, wetness and haze while preserving scene identity. For final video synthesis, we compare a naive prompt (“a heavy snowy driving scene with snowflakes.”) with our captioned prompt. Without detailed guidance, the model is prone to hallucinations because injected particles alter the depth map and can contradict the model’s structural prior. The detailed prompt provides stronger semantic constraints, helping the model interpret these perturbations as weather effects, yielding the desired, visually consistent results across frames.

\noindent\textbf{Geometry grounding.} As shown in ~\cref{fig:ablation_geometry}, we ablate the influence of gravity direction matching and camera-consistent projection on particle motion in geometry grounding. For clarity, we visualize particle trajectories using the rendered particle maps. Without gravity direction alignment, snowflake motion is dominated by the canonical axis, yielding vertical falling; with alignment, particles follow the real scene gravity direction estimated from the fused point map, producing consistent trajectories under tilted cameras. To demonstrate the effectiveness of camera projection, we disable physics-driven motion and keep only motion induced by the camera. In a forward-driving sequence, particles should exhibit strong parallax and appear to approach the viewer; without camera projection, particles remain stationary relative to the image, whereas with projection, they move coherently with ego-motion, yielding geometry-consistent particle motion.



\begin{figure}[t!]
  \centering
   \includegraphics[width=\linewidth]{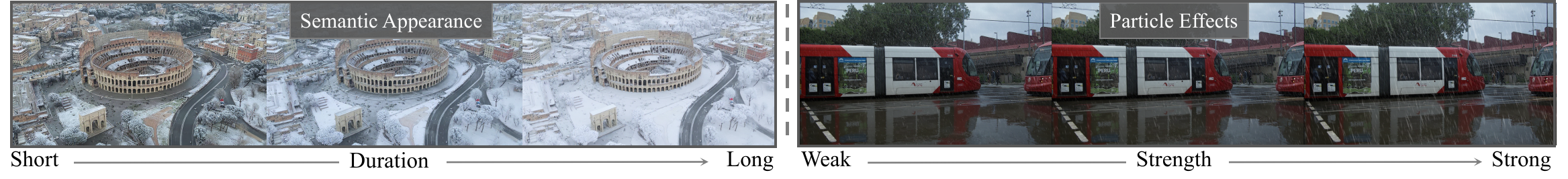}\caption{\textbf{Examples of controllable weather effects.} (Left) User inputs specify the weather duration category (short/medium/long) to control overall semantic appearance. (Right) Simulation parameters adjust particle shape and motion to reflect wind strength (weak to strong), enabling fine-grained particle control.} 
   \label{fig:control}
\end{figure}

\begin{figure}[t!]
\centering

\begin{minipage}[t]{0.48\linewidth}
\vspace{0pt}
\centering
\small
\resizebox{\linewidth}{!}{%
\begin{tabular}{lcccc}
\toprule
Method & \multicolumn{2}{c}{\textbf{ACDC}} & \multicolumn{2}{c}{\textbf{MUSES}} \\
\cmidrule(lr){2-3} \cmidrule(lr){4-5}
& Snow & Rain & Snow & Rain \\
\midrule
\textbf{DAFormer} & 50.9 & 50.8 & 41.4 & 33.8 \\
\textbf{HRDA}     & 50.0 & 51.6 & 40.4 & 35.4 \\
\midrule
\textbf{DAFormer+ours} &
56.3~\textcolor{mygreen}{+5.4} &
56.6~\textcolor{mygreen}{+5.8} &
47.2~\textcolor{mygreen}{+5.8} &
41.5~\textcolor{mygreen}{+7.7} \\
\textbf{HRDA+ours} &
60.4~\textcolor{mygreen}{+10.4} &
66.1~\textcolor{mygreen}{+14.5} &
50.4~\textcolor{mygreen}{+10.0} &
44.4~\textcolor{mygreen}{+9.0} \\
\bottomrule
\end{tabular}}
\captionof{table}{Quantitative segmentation results for baselines and models augmented with our data. Improvements over the baseline are shown in green.}
\label{tab:seg}
\end{minipage}
\hfill
\begin{minipage}[t]{0.50\linewidth}
\vspace{0pt}
\centering
\includegraphics[width=\linewidth]{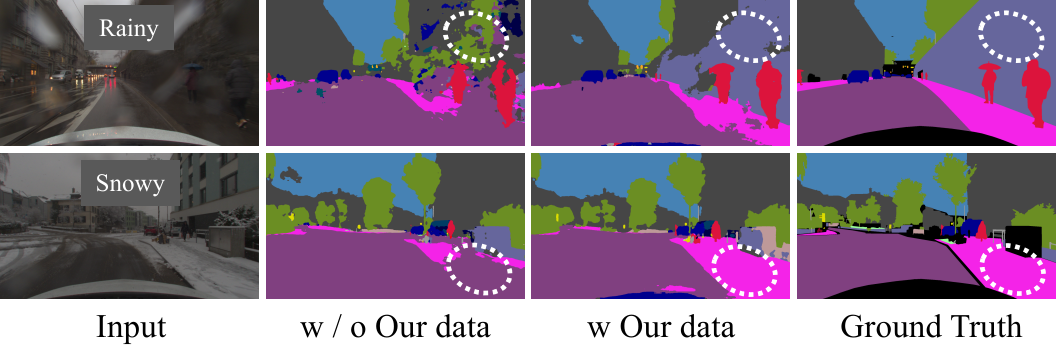}
\captionof{figure}{Qualitative segmentation results comparison under adverse weather: w/ vs. w/o our augmentation. Improved regions are highlighted in white circles.}
\label{fig:seg}
\end{minipage}
\vspace{-3mm}
\end{figure}

\subsection{Applications}
By leveraging the structured conditioning, our design enables flexible and interpretable control over both semantic appearance and particle effects, as shown in Fig.~\ref{fig:control}. This disentangled control allows users to flexibly synthesize diverse weather conditions for the same scene, ranging from subtle environment changes to extreme conditions with dense particles. We further demonstrate the utility of our synthesized data for augmenting semantic segmentation in the domain generalization setting. When training DAFormer\cite{hoyer2022daformer} and HRDA\cite{hoyer2022hrda} with our generated data, we observe consistent improvements across evaluation settings, boosting mIoU by at least 5.4\% and up to 14.5\% (\cref{tab:seg}). Qualitatively, our augmentation mitigates prediction errors of the HRDA\cite{hoyer2022hrda} baseline on MUSES\cite{brodermann2024muses}, yielding cleaner, more complete segmentation masks (\cref{fig:seg}). These results indicate that our synthesized weather data provides a valuable supplement for training perception models robust to adverse weather.

\section{Limitations}
Our method builds upon several upstream modules, including first-frame anchoring, geometry estimation, gravity alignment, and particle projection. Errors at any of these stages may propagate directly into the final video. For instance, the geometry grounding relies on recovering a dominant ground plane to estimate gravity, which can be fragile in scenes that lack clear ground structure or involve challenging viewpoints. Moreover, the achievable weather particle effects are inherently bounded by the pretrained video diffusion model, occasionally leading to failed particle generation. Future work may address these limitations by jointly optimizing the upstream modules to mitigate cascading errors and incorporating stronger or domain-adapted diffusion priors to improve the fidelity and controllability of particle effects.
\section{Conclusion}
In this work, we propose a semantic-aware, physics-informed, and geometry-grounded framework for controllable weather synthesis in real-world videos. We factorize the synthesis into a structured tri-prior interface comprising semantics, dynamics, and geometry, enabling precise control over both global atmospheric appearance and fine-grained particle motion. This design produces diverse and physically plausible weather effects across varied scenes, effectively eliciting weather priors that are often underexpressed with text-only prompting. Furthermore, the resulting high-fidelity synthetic data improves downstream robustness, yielding up to 14.5\% mIoU improvement in semantic segmentation under extreme weather conditions. It highlights the effectiveness of our method as a scalable engine for generating diverse, safety-critical corner cases, enabling controllable synthesis of rare adverse weather conditions that are difficult to capture in the real world, thereby improving downstream perception robustness for resilient autonomous systems.

\section*{Acknowledgment}
This work was funded by the National Natural Science Foundation of China (Nos.~62572166 and 62402157) and the Fundamental Research Funds for the Central Universities (No.~JZ2025HGTB0219). It was also partially supported by the Ministry of Education and Science of Bulgaria (support for INSAIT, part of the Bulgarian National Roadmap for Research Infrastructure).

\bibliographystyle{plainnat}
\bibliography{references}

\end{document}